\documentclass[runningheads]{llncs}
\usepackage[T1]{fontenc}
\usepackage{graphicx}
\usepackage{booktabs}
\usepackage[misc]{ifsym}
\newcommand{\corr}{(\Letter)}
\usepackage{mwe}

\usepackage{hyperref}

\usepackage{amsmath}
\usepackage{amssymb}
\usepackage{mathtools}
\usepackage{algorithm}
\usepackage{algorithmic}

\usepackage{tikz}
\usetikzlibrary{calc}
\usepackage{subcaption} 
\usepackage{graphicx}

\usepackage{multirow} 
\usepackage{hyperref}

\usepackage{xcolor}

\begin{document}

\title{Gradient Similarity Surgery\\ in Multi-Task Deep Learning}

\titlerunning{Multi-Task Optimisation Similarity Based}

\author{
Thomas Borsani\inst{1} \corr  \and 
Andrea Rosani\inst{1} \and \\
Giuseppe Nicosia\inst{2} \and
Giuseppe Di Fatta\inst{1}
}

\institute{
Free University of Bozen-Bolzano, Bozen-Bolzano, Italy \\
\email{\{tborsani, Andrea.Rosani, Giuseppe.DiFatta\}@unibz.it} \and 
University of Catania, Catania, Italy \\ \email{nicosia@dmi.unict.it}
}

\authorrunning{T. Borsani et al.}

\maketitle

\begin{abstract}
The multi-task learning ($MTL$) paradigm aims to simultaneously learn multiple tasks within a single model capturing higher-level, more general hidden patterns that are shared by the tasks. 
In deep learning, a significant challenge in the backpropagation training process is the design of advanced optimisers to improve the convergence speed and stability of the gradient descent learning rule. 
In particular, in multi-task deep learning ($MTDL$) the multitude of tasks may generate potentially {\it conflicting gradients} that would hinder the concurrent convergence of the diverse loss functions. 
This challenge arises when the gradients of the task objectives have either different magnitudes or opposite directions, causing one or a few to dominate or to interfere with each other, thus degrading the training process.
Gradient surgery methods address the problem explicitly dealing with conflicting gradients by adjusting the overall gradient trajectory.
This work introduces a novel gradient surgery method, the Similarity-Aware Momentum Gradient Surgery ({\sc SAM-GS}), which provides an effective and scalable approach based on a gradient magnitude similarity measure to guide the optimisation process.
The {\sc SAM-GS} surgery adopts gradient equalisation and modulation of the first-order momentum.
A series of experimental tests have shown the effectiveness of {\sc SAM-GS} on synthetic problems and $MTL$ benchmarks. Gradient magnitude similarity plays a crucial role in {\it regularising gradient aggregation} in $MTDL$ for the optimisation of the learning process. Code is available at \href{https://unibzmlgroup.github.io/SAMGS/}{https://unibzmlgroup.github.io/SAMGS/}

\keywords{Multi-Task Deep Learning 
\and Gradient Descent Optimisation 
\and Gradient Surgery 
\and Gradient Aggregation 
\and Conflicting Gradients 
}
\end{abstract}

\section{Introduction}
In the multi-task learning ($MTL$) paradigm \cite{Caruana1997} a model is trained on multiple tasks simultaneously, leveraging a shared internal representation to improve generalisation and efficiency. 
While training a model for a single task leverages on patterns in the data, training on multiple tasks also leverages on patterns in the tasks. 
$MTL$ exploits task similarities to enhance performance, particularly when tasks share some underlying features.
Utilising a shared representation for many tasks allows to improve model generalisation by capturing features that are more resilient to noise compared to a single-task approach. 
This concurrent learning process acts as a regularisation mechanism, reducing bias and strengthening the robustness of the model. 
Additionally, this approach is advantageous when data availability is particularly heterogeneous across tasks, as it enables the aggregation of data from many tasks to improve overall learning.
Moreover, $MTL$ can lead to a reduction in computational costs, training and inference time, but this depends on the specific implementation and task relationships.

The $MTL$ paradigm has been successfully applied to many problems across various domains, including Natural Language Processing \cite{Ruder2017,Wang2018}, Computer Vision \cite{Choi2024MultitaskLF,Tian2024}, Healthcare and Medical Imaging \cite{Hao2022,Sangwook2023}, Fraud Detection and Finance \cite{Ong2023}.
These applications demonstrate how $MTL$ can improve generalisation, reduce data requirements, and enhance model efficiency across diverse real-world problems.
Nevertheless, there are challenges to effectively training $MTL$ models, particularly in selecting and combining tasks, as different tasks may not always align seamlessly to produce better solutions \cite{Standley2020ICML}.

Recent research has evidenced that one of the primary challenges for the optimisation of $MTL$ models is the aggregation of the different gradients associated to the task-specific loss functions \cite{Yu2020NeurIPS_pcgrad}. 
Typically, the task gradients are aggregated using the arithmetic mean.
Indeed, it has been shown that this approach can lead to suboptimal solutions \cite{Standley2020ICML,Yu2020NeurIPS_pcgrad}.
The underlying cause have been identified in the challenges arising from the aggregation of conflicting task gradients, i.e. gradients with opposite directions (\textit{angle-based conflicting gradients}) and gradients dominating the aggregation (\textit{magnitude conflicting gradients}) \cite{Yu2020NeurIPS_pcgrad}.

Current solutions to address the problem of conflicting gradients can be categorised into three sub-groups. 
\textit{Task Similarity} methods focus on the selection of tasks that do not cause gradient conflicts \cite{Fifty2022NeurIPS,Zamir2018CVPR}. 
\textit{Loss Balancing} methods focus on static or dynamic weighting algorithms to weight the different loss functions \cite{Liu2023NeurIPS_famo,Chen2018ICML_gradnorm}, and \textit{Gradient Surgery} methods seek to mitigate gradient conflicts by applying heuristics that modify the gradient descent learning rule to reduce their impact \cite{Yu2020NeurIPS_pcgrad,Liu2021NeurIPS_cagrad,Navon2022_nashmtl}.

However, methods of \textit{Task Similarity} tend to be computationally inefficient and limit $MTL$ applicability, serving primarily to avoid the problem rather than addressing it to optimise the potential benefits offered by $MTL$ models.
\textit{Loss Balancing} methods, while effective and more efficient than task similarity methods in addressing the problem \cite{Liu2023NeurIPS_famo}, still ignore its underlying causes. 
\textit{Gradient Surgery} methods tackle gradient conflicts directly and have been shown to be among the most effective strategies to optimise $MTL$ models \cite{Liu2021ICLR_imtl,Navon2022_nashmtl,Senushkin2023_alighedmtl}. 
Most of these methods, however, apply the procedure indiscriminately, overlooking the proper identification of gradient conflicts, which can lead to a deterioration of the original gradient-based learning process.
Additionally, some of these methodologies excessively level out the relative contributions of the tasks to the overall gradient, and inevitably miss out the inherent advantage of $MTL$, where tasks may provide complementary contributions in the shared representation. 

To address the issue of conflicting gradients while accounting for the varying nature of task loss functions, we introduce a novel gradient surgery method, the Similarity-Aware Momentum Gradient Surgery (SAM-GS).
This method dynamically adapts the gradient descent optimisation process based on the task gradient magnitude similarity.
The proposed approach applies a conservative learning when gradients are dissimilar and accelerates learning when they exhibit high similarity. 
SAM-GS integrates gradient equalisation within conflicting scenarios and incorporates a gradient momentum, whose influence is adaptively modulated based on the task gradient similarity.
Comparative experimental results demonstrate that this adaptive strategy enhances stability and efficiency in learning dynamics, yielding superior performance across diverse $MTL$ benchmarks.

Key contributions of the proposed SAM-GS method are as follows:
\begin{itemize}
    \item \textit{SAM-GS Optimisation}: Introduction of a gradient similarity measure to selectively adjust gradient magnitudes, enhancing the learning process.
    \item \textit{Momentum-Based Regularisation}: Integration of gradient momentum into gradient surgery, introducing a new regularisation for conflicting gradients, improving the optimisation dynamics.
    \item \textit{Empirical Validation}: Analysis on synthetic problems and evaluation on four standard $MTL$ benchmarks, achieving comparable or improving state-of-the-art (SOTA) performance over existing methods.
\end{itemize}
\noindent    
The remainder of the paper is organised as follows. 
In Section \ref{Sec:MTLOptimisation}, we present the problem of conflicting gradients in $MTDL$ and the solution offered by gradient surgery methods.
In Section \ref{Sec:Related Works},  we discuss related work in terms of the three approaches, task similarity, load balancing and gradient surgery, to compare and contrast them.
In Section \ref{Sec:SAMGS}, the proposed SAM-GS method is introduced and its main algorithm described.
In Section \ref{Sec:Experiments and Comparisons}, we present an experimental and comparative analysis of the proposed method with respect to other gradient surgery methods.
Section \ref{Sec:Conclusions} provides the main conclusions and indicates some areas of improvement.

\section{$MTL$ Optimisation}\label{Sec:MTLOptimisation}
In this section, we introduce the definition of the multi-task learning paradigm, discuss the specific challenge referred to as \emph{conflicting gradients} in deep learning models, and provide an overview of gradient surgery methods.

The $MTL$ paradigm aims to optimise a single model $\theta \in \mathbb{R}^m$ for $K \geq 2$ numbers of tasks simultaneously. In general, the objective is to minimise the sum of the task-specific loss functions $\mathcal{L}_i (\theta): \mathbb{R}^m \rightarrow \mathbb{R}_{+}$

\begin{equation}
\label{formula:MTL model definition}
     \arg\min_{\theta \in \mathbb{R}^m} \biggl\{ \mathcal{L}_{mtl}(\theta):=\sum_{i=1}^K \mathcal{L}_i (\theta) \biggl\}
\end{equation}

The training of a $MTL$ model through direct optimisation of the Equation \eqref{formula:MTL model definition} may yield to sub-optimal solutions, characterised by under-optimised tasks \cite{Yu2020NeurIPS_pcgrad}. 
More specifically, $MTL$ can be framed as a multi-objective optimisation problem \cite{DiFatta2020}, where optimising Equation \eqref{formula:MTL model definition} may result in solutions that are not Pareto-efficient.

In deep network models the literature has identified gradient conflicts as one of the primary causes of this sub-optimisation issue \cite{Yu2020NeurIPS_pcgrad}.

\subsection{Conflicting Gradients in $MTDL$} 
 
In training deep learning models on multiple tasks simultaneously, the issue of conflicting gradients arises when different tasks produce gradients that interfere with each other, leading to inefficient or suboptimal learning. This detrimental interference hinders the performance of the model across tasks.

Two main types of conflicting gradients can be identified, respectively, caused by the relative direction of the task gradient vectors and by their different magnitudes.

\subsubsection{Angle-Based Gradient Conflict.} 
Let \( g_i, g_j \in \mathbb{R}^d \) be the gradient vectors associated with two different tasks \( i \) and \( j \). We define an \textit{angle-based gradient conflict} as occurring when the angle \( \phi_{ij} \), in Equation \eqref{formula:cosine similarity}, between them is greater than $90°$, which corresponds to a negative cosine similarity. In this situation, the vector sum reduces the net effective learning step, slowing convergence \cite{Yu2020NeurIPS_pcgrad}.

\begin{equation}
\label{formula:cosine similarity}
    \cos(\phi_{ij}) = \frac{g_i \cdot g_j}{\|g_i\| \|g_j\|} < 0.
\end{equation}

In this scenario, the least critical case occurs when the gradients from different tasks are nearly orthogonal to each other. 
This still results in inefficient learning since updates get diluted rather than reinforcing progress in the common direction. 
The most critical case arises when the gradients from different tasks are perfectly opposite to each other, resulting in a zero vector and effectively preventing learning.

\subsubsection{Magnitude Gradient Conflict.}
Let \( g_i, g_j \in \mathbb{R}^d \) be the gradient vectors associated with two different tasks \( i \) and \( j \). 
We quantify the \textit{magnitude gradient conflict} by means of the magnitude similarity defined in Equation \eqref{formula:magnitude similarity}: 

\begin{equation}
\label{formula:magnitude similarity}
\psi(g_i, g_j) = \frac{2 \|g_i\|_2 \|g_j\|_2}{\|g_i\|_2 ^2 + \|g_j\|_2 ^2}.
\end{equation}

A magnitude gradient conflict occurs when the gradients associated with different tasks have significantly varying magnitudes. 
This imbalance can cause the model to prioritise certain tasks over others, leading to suboptimal performance.

In contrast to the \textit{angle-based gradient conflict}, where it is clearly defined when two gradients are in conflict (i.e., negative cosine similarity), the detection of magnitude-based gradient conflicts is less straightforward. Dissimilarities in task gradient magnitudes may not be due to actual conflicts but to the lack of loss normalisation or to local topological differences in loss functions across the tasks.

\subsection{Gradient Surgery Methods} 
Gradient surgery methods provide a heuristic aggregation function over the task gradient vectors to compute the overall gradient driving the weight update rule. The surgery function is aimed at optimising all tasks effectively by limiting the effect of gradient conflicts.

We introduce a generic task gradient aggregation function, which determines how gradients from different tasks are combined according to the surgery method.
The gradient of the total loss with respect to the weight matrix $\theta$ at layer is \( l \) is:
\(\nabla_{\theta^{(l)}} \mathcal{L} = s\left( \nabla_{\theta^{(l)}} \mathcal{L}_1, \nabla_{\theta^{(l)}} \mathcal{L}_2, \dots, \nabla_{\theta^{(l)}} \mathcal{L}_K \right),\)
where \( s(\cdot) \) is a task gradient aggregation function that determines how the individual task gradients contribute to the overall optimisation.

\section{Related Works}\label{Sec:Related Works}
Existing solutions to deal with gradients conflicts can been categorised in three main groups, as follows.

\paragraph{Task Similarity.} 
The optimisation via Task Similarity methods aims to group tasks that can be learned synergistically, thereby improving overall model performance.
It is also possible that the best solution does not involve using one $MTL$ model to solve $K$ tasks, but rather employing $K$ single-task models, which may lead to better outcomes \cite{Fifty2022NeurIPS,Zamir2018CVPR,Shi2023_recon,Standley2020ICML,Shen2021NeurIPS}.
In this approach, gradient conflicts are avoided by selecting a suitable combination of tasks that do not present conflicts.

\paragraph{Loss Balancing.}
Loss Balancing methods relies on weighting the different loss functions of the tasks involved in the combination. 
Various methodologies have been proposed to determine the optimal weights for different tasks. 
The method \textbf{UW} \cite{Kendall2018CVPR_uw} leverages the homoscedastic uncertainty of each task to determine the weights , while \textbf{DWA} \cite{Liu2019CVPR_dwa} utilises rate of change of task-specific loss functions. 
\textbf{GradNorm} \cite{Chen2018ICML_gradnorm} modulates weights based on the magnitude of the gradient. 
In contrast to these approaches, \textbf{RLW} \cite{Lin2022TRML_RGW} assigns random weights.
Additionally, \textbf{FAMO} \cite{Liu2023NeurIPS_famo} learns the weights based on the quality of the loss updates.
These methods mitigates gradient conflicts by preventing any single task from dominating the training process.

\paragraph{Gradient Surgery.} 
These methods aim to enhance convergence in $MTL$ by appropriately weighting the gradient components of different tasks. 
They focus on introducing heuristics to adjust the combination of gradient vectors, thereby influencing the optimisation process dynamics to resolve the conflicts and guiding the model more effectively through the loss landscape. 
Approaches like \textbf{Nash-MTL} \cite{Navon2022_nashmtl} utilise game theory concepts, particularly the Nash Bargaining Solution, to equilibrate task gradients.
Instead, \textbf{MGDA} \cite{DESIDERI2012_mgda,Dong2015_mgda2} for $MTL$ seeks a direction that minimises all objectives simultaneously, in line with the multi-objective Karush–Kuhn–Tucker (KKT) \cite{Harold_1951_KKT} conditions.
These methods are computationally intensive but have proven to be effective.

Alternative approaches aim to mitigate gradient conflicts. \textbf{GDOD} \cite{Dong2023} decomposes task gradients into shared and conflicting components, updating only the shared ones. \textbf{PCGrad} \cite{Yu2020NeurIPS_pcgrad} reduces conflicts among task gradients by decorrelating them while 
\textbf{CAGrad} \cite{Liu2021NeurIPS_cagrad} seeks a conflict-averse gradient path to minimise task interference. 
\textbf{GradDrop} \cite{Chen2020NeurIPS_graddrop} ensures consistency in gradient signs across tasks.
In addition, \textbf{IMTL} \cite{Liu2021ICLR_imtl} identifies a gradient path in which cosine similarities among task gradients remain consistent, and \textbf{Aligned-MTL} \cite{Senushkin2023_alighedmtl} mitigates conflicts by aligning the principal components of the gradient matrix. 
These methods compete effectively with the more complex Nash-MTL \cite{Navon2022_nashmtl} showing better performance maintaining low computational overhead.

\section{Similarity-Aware Momentum Gradient Surgery}\label{Sec:SAMGS}

Similarity-Aware Momentum Gradient Surgery (SAM-GS) is a gradient surgery method that leverages a measure of the magnitude similarity of the task gradients to detect and address conflicts during the learning process.

Here, we first present the intuition behind the approach with an example with four scenarios, and then we introduce the SAM-GS algorithm. 

The proposed approach focuses solely on \textit{magnitude gradient conflicts}, which are arguably critical to effective $MTDL$ optimisation. 

\textit{Angle-based gradient conflicts} are intentionally disregarded, as they only impact convergence speed.

The core difficulty of $MTDL$, compared to $STL$, stems from the presence of \textit{magnitude gradient conflicts}, which are unique to $MTDL$ and the primary source of task-specific conflicts \cite{Elich_2025_gcpr}. 
In contrast, \textit{Angle-based gradient conflicts} are more characteristic of inter-sample variation typically address with mini-batch gradient descent.

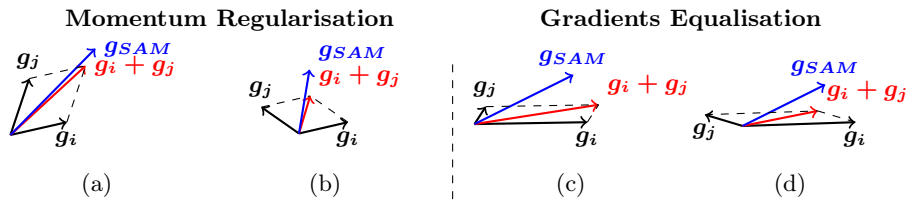
\begin{figure}[h!t]
\centering
\makebox[0\textwidth][l]{\textbf{Momentum Regularisation}}  
    \hspace{0.5\textwidth}  
    \textbf{Gradients Equalisation}\\
    \begin{subfigure}{0.23\textwidth}
        \centering
        
        \begin{tikzpicture}
            
            \coordinate (gi) at (.75,0.17);     
            \coordinate (gj) at (.25,.75);     

            \draw[->, thick] (0,0) -- (gi) node[anchor=north] {$\vec{g_i}$};
            \draw[->, thick] (0,0) -- (gj) node[anchor=south] {$\vec{g_j}$};

            \coordinate (sum) at ({.75 + (.25)}, {0.17 + .75});

            \draw[->, thick, red] (0,0) -- (sum) node[anchor=west] {$\vec{g_i+g_j}$};

            \coordinate (sumsam) at ({.75 + (.25)+0.13}, {0.17 + .75+0.23});
            \draw[->, thick, blue] (0,0) -- (sumsam) node[anchor=west] {$\vec{g_{SAM}}$};

            \draw[dashed] (.75,0.17) -- (sum);
            \draw[dashed] (.25,.75) -- (sum);
        \end{tikzpicture}
        \caption{}
        \label{fig:vector_angle_good(a)}
    \end{subfigure}
    \begin{subfigure}{0.25\textwidth}
        \centering
        \begin{tikzpicture}
            
            \coordinate (gi) at (.65,0.15);     
            \coordinate (gj) at (-.5,.35);     

            \draw[->, thick] (0,0) -- (gi) node[anchor=north] {$\vec{g_i}$};
            \draw[->, thick] (0,0) -- (gj) node[anchor=south] {$\vec{g_j}$};

            \coordinate (sum) at ({.65 + (-.5)}, {0.15 + .35});

            \draw[->, thick, red] (0,0) -- (sum) node[anchor=south west] {$\vec{g_i+g_j}$};
            
            \coordinate (sumsam) at ({.65 + (-.5)-0.01}, {0.15 + .35+0.35});
            \draw[->, thick, blue] (0,0) -- (sumsam) node[anchor=south west] {$\vec{g_{SAM}}$};

            \draw[dashed] (.65,0.15) -- (sum);
            \draw[dashed] (-.5,.35) -- (sum);
        \end{tikzpicture}
        \caption{}
        \label{fig:vector_angle_angle(b)}
    \end{subfigure}
    \begin{subfigure}{0.01\textwidth}
        \centering
        \begin{tikzpicture}
            \draw[dashed] (0,-1) -- (0,.8);
        \end{tikzpicture}
    \end{subfigure}
    \begin{subfigure}{0.23\textwidth}
        \centering

        \begin{tikzpicture}
            
            \draw[->, thick] (0,0) -- (1.5,.025) node[anchor=north] {$\vec{g_i}$};

            \draw[->, thick] (0,0) -- (0.15,.232) node[anchor=south] {$\vec{g_j}$};

            \coordinate (sum) at ({0.15 + (1.5)}, {0.025 + .232});

            \draw[->, thick, red] (0,0) -- (sum) node[anchor=south west] {$\vec{g_i+g_j}$};

            \coordinate (sumsam) at ( {1.3155},{.6542});
            \draw[->, thick, blue] (0,0) -- (sumsam) node[anchor=south] {$\vec{g_{SAM}}$};

            \draw[dashed] (1.5,.025) -- (sum);
            \draw[dashed] (0.15,.232) -- (sum);
        \end{tikzpicture}
        \caption{}
        \label{fig:vector_angle_mang(c)}
    \end{subfigure}
    \begin{subfigure}{0.23\textwidth}
        \centering
        \begin{tikzpicture}
            
            \coordinate (gi) at (1.5,0.05);     
            \coordinate (gj) at (-.5,.15);     

            \draw[->, thick] (0,0) -- (gi) node[anchor=north] {$\vec{g_i}$};
            \draw[->, thick] (0,0) -- (gj) node[anchor=north] {$\vec{g_j}$};

            \coordinate (sum) at ({1.5 + (-.5)}, {0.15 + .05});

            \draw[->, thick, red] (0,0) -- (sum) node[anchor=south west] {$\vec{g_i+g_j}$};

            \coordinate (sumsam) at ({1.1000}, {0.5500});
            \draw[->, thick, blue] (0,0) -- (sumsam) node[anchor=south] {$\vec{g_{SAM}}$};

            \draw[dashed] (1.5,0.05) -- (sum);
            \draw[dashed] (-.5,.15) -- (sum);
        \end{tikzpicture}
        \caption{}
        \label{fig:vector_angle_and_mang(d)}
    \end{subfigure}

    \caption{Illustration of four scenarios for two task gradients, $\vec{g_i}$ and $\vec{g_j}$, the standard overall gradient is denoted as $\vec{g_i+g_j}$ and the overall gradient of SAM-GS is denote as $\vec{g_{SAM}}$. (a) Ideal case: Gradients have similar magnitudes, and the angle between them is less than $90^\circ$, indicating no conflict. (b) \textit{angle-based gradient conflict}: The angle between gradients exceeds $90^\circ$, diminishing the effectiveness of their combination. (c) \textit{magnitude-based gradient conflict}: One gradient dominates, leading to an imbalanced gradient update. (d) Both conflicts: A combination of angle- and magnitude-based gradient conflicts, where both the directional misalignment and magnitude disparity hinder effective gradient aggregation.}
    \label{fig:vector_angle}
\end{figure}

Let us consider why \textit{angle-based gradient conflicts} can slow the convergence of the learning process while \textit{magnitude gradient conflicts} can significantly hinder the overall optimisation preventing the convergence of some tasks.
When adding two vectors ${g}_i$ and ${g}_j$ of similar magnitude ($|g_i| \simeq |g_j|$) at an angle $\alpha$ greater than $90^{\circ}$, the magnitude of the sum is reduced by a factor proportional to $cos(\alpha)$ compared to adding them when they are collinear, as shown in Figure \ref{fig:vector_angle_good(a)} and \ref{fig:vector_angle_angle(b)}. In the worst case, when $\alpha=180$, the two vectors are in exactly opposite directions, and their magnitudes cancel out. However, this extreme case is rather unlikely. Although reduced in magnitude, the vector sum still contains useful information about the direction of optimisation for the gradient descent algorithm. Hence, to enhance the magnitude of the resulting sum vector by means of the momentum with no need to detect this type of conflict explicitly.

However, when one of the task gradients is overly greater than the others ($|g_i| \gg |g_j|$) the overall sum of the gradients will result in a direction dominated by that single vector. This case can be quite detrimental as only one task will benefit from the learning process, as shown in Figure \ref{fig:vector_angle_mang(c)}.
In this case, we introduce a conflict detection mechanism and a procedure to equalise the task gradients before their aggregation.

As illustrated in Figure \ref{fig:vector_angle}, therefore, \textit{magnitude gradient conflicts} have the potential to steer the optimisation process away from a fair convergence of all tasks, while \textit{angle-based gradient conflicts} only influence the pace of convergence.

For this reason, SAM-GS ignores \textit{angle-based gradient conflicts} and introduces two mechanisms: momentum regularisation and gradients equalisation.
In particular, the momentum is modulated by the magnitude similarity, and the gradients equalisation is triggered by the detection of \textit{magnitude gradient conflicts} by means of the magnitude similarity.

In cases where task gradients exhibit significantly different magnitudes, our approach equalises their magnitudes to compute a balanced direction not dominated by one task. The resulting sum vector is then scaled by the average magnitude to prevent the occurrence of near-zero gradients.

SAM-GS follows a general structure that is similar to ADABelief \cite{Zhuang2020NeurIPS}.
SAM-GS is specifically designed for multi-gradient optimisation, while ADABelief is applied to a single gradient ($STL$).
ADABelief adopts a regularisation of the momentum that is based on the gradient, whereas SAM-GS applies a regularisation technique based on a gradient similarity measure. 

Accordingly, SAM-GS is presented in Algorithm \ref{algorithm:sam-gs}, where $\gamma$ is a learnable hyperparameter to set the threshold on the gradient similarity to detect \textit{magnitude gradient conflicts}.
Let the model parameter vector at step $t$ be represented by $\theta_t$, it follows that for each of the $K\geq2$ tasks, there exist a differentiable loss function,  \(\{l_i\}_{i=1} ^K\). 
Consequently, for each task, the gradients \(g_k = \nabla_{\theta} \mathcal{L}_k\) can be computed.
The average magnitude similarity of the gradients, denoted as \( \Psi_t \), is computed from the gradient magnitude similarities of Equation \eqref{formula:magnitude similarity}.  
Furthermore, we indicate the momentum with $m_{k,t}$, which is the exponential moving average (EMA) of $g_{k,t}$, and with $h_t$ the EMA of $(1-\Psi_t)^2$ (similarity momentum coefficient) with $\beta_1$ and $\beta_2$ the smoothing parameters and $\hat{.}$ represents the bias-corrected value of the respective quantity.

\begin{algorithm}[h!t]
   \caption{Similarity-Aware Momentum Gradient Surgery}
   \label{algorithm:sam-gs}
\begin{algorithmic}
    \STATE {\bfseries Hyperparameters:} $\beta_1 \leftarrow 0.9, \beta_2 \leftarrow 0.99, \gamma \leftarrow 0.1$ 
    \STATE {\bfseries Initialise:} $\theta_0, m_0 \leftarrow 0 , h_0 \leftarrow 0, t \leftarrow 0, \epsilon \leftarrow 1e-8 $
    \REPEAT
    \STATE $t \leftarrow t + 1$
    \STATE $\vec{g}_k \leftarrow \nabla_{\theta_t} \mathcal{L}_k, \forall k $

    \STATE $\Psi = \frac{1}{K^2} \sum_{i,j} \psi(g_i, g_j)$ 
    \STATE $\vec{m}_{k,t} \leftarrow \beta_1 \vec{m}_{k,t-1} + (1 - \beta_1) \vec{g}_k, \forall k $
    \STATE $h_t \leftarrow \beta_2 h_{t-1} + (1 - \beta_2)(1 -\Psi)^2 +\epsilon$
    \STATE $\widehat{\vec{m}}_{k,t} \leftarrow \frac{\vec{m}_{k,t}}{1-\beta_1 ^t}$,  $\widehat{h}_t \leftarrow \frac{h_t}{1-\beta_2 ^t}$
    \IF{$\Psi < \gamma$}
    
    \STATE $\vec{w}_k = \frac{ \overline{ \|\vec{g}_{k}\|}_2}{\|\vec{g}_{k}\|_2} \vec{g}_k, \forall k$
    \ELSE 
    \STATE $ \vec{w}_k = \frac{|\widehat{\vec{m}}_{k,t}|} {\sqrt{\widehat{h}_t}+\epsilon}, \forall k$
    \ENDIF
    
    \STATE {\bfseries Update:} $\theta_t = \theta_{t-1} - \alpha \sum_{k=1}^{K} \vec{w}_k \odot \vec{g}_k$
   \UNTIL{convergence}
\end{algorithmic}
\end{algorithm}

The proposed SAM-GS approach mitigates gradient dominance by adopting cautious updates with smaller step sizes. 
Conversely, when gradients are well-balanced, it leverages the momentum to accelerate learning and compensate for prior conservative updates.
$h_t$ acts as a regularisation term, where, if the gradients are dissimilar, the momentum is trusted less. 
Conversely, when the gradients exhibit good magnitude similarity, the momentum retains its full potential. 
The parameter $\gamma$ plays a crucial role in determining the threshold at which gradients are considered well-balanced. 
We provide an ablation study on this parameter in section \ref{section:ablation}.

\section{Computational Experiments and Comparisons} \label{Sec:Experiments and Comparisons}
We conduct a series of experiments to empirically demonstrate the effectiveness of SAM-GS compared to other methods on synthetic problems and on common multi-task supervised benchmarks.
Two variants of a synthetic problem based on two parameters are used to highlight the effect of gradient conflicts and how different methods fair under such conditions.
The benchmarks based on real-world problems allow a comparative performance analysis of the proposed method against many state-of-the-art optimisation methods for $MTL$. An ablation study of SAM-GS hyperparameter $\gamma$ allows to investigate its impact on the performance of the method.
In the following, each experimental setup is described and the results are presented.

\subsection{Synthetic Problem}
To illustrate the gradient surgery problem in a simplified setting, we adopt the 2D multi-task optimisation problem proposed in Nash-MTL \cite{Navon2022_nashmtl}. 
This problem provides a controlled environment for the study of conflicting gradient across tasks, highlighting the challenges of multi-task optimisation.
In addition, we introduce a novel variant of that problem with a similar loss landscape structure, featuring two global minima, providing a different problem setting to analyse the impact of multiple optima on optimisation dynamics.

\subsubsection{Two-task problem with one global optimum.}

The synthetic problem proposed in \cite{Navon2022_nashmtl} provides a useful toy problem to investigate and visualise the behaviour of multi-task optimisation methods in a complex yet comprehensible loss landscape. 
The problem consists of two loss functions with two parameters, and the objective is to minimise both using an $MTL$ optimisation approach; a detailed formulation is reported in \cite{Liu2021NeurIPS_cagrad}.

\begin{figure}[h!t]
    \centering
    \includegraphics[width=\linewidth]{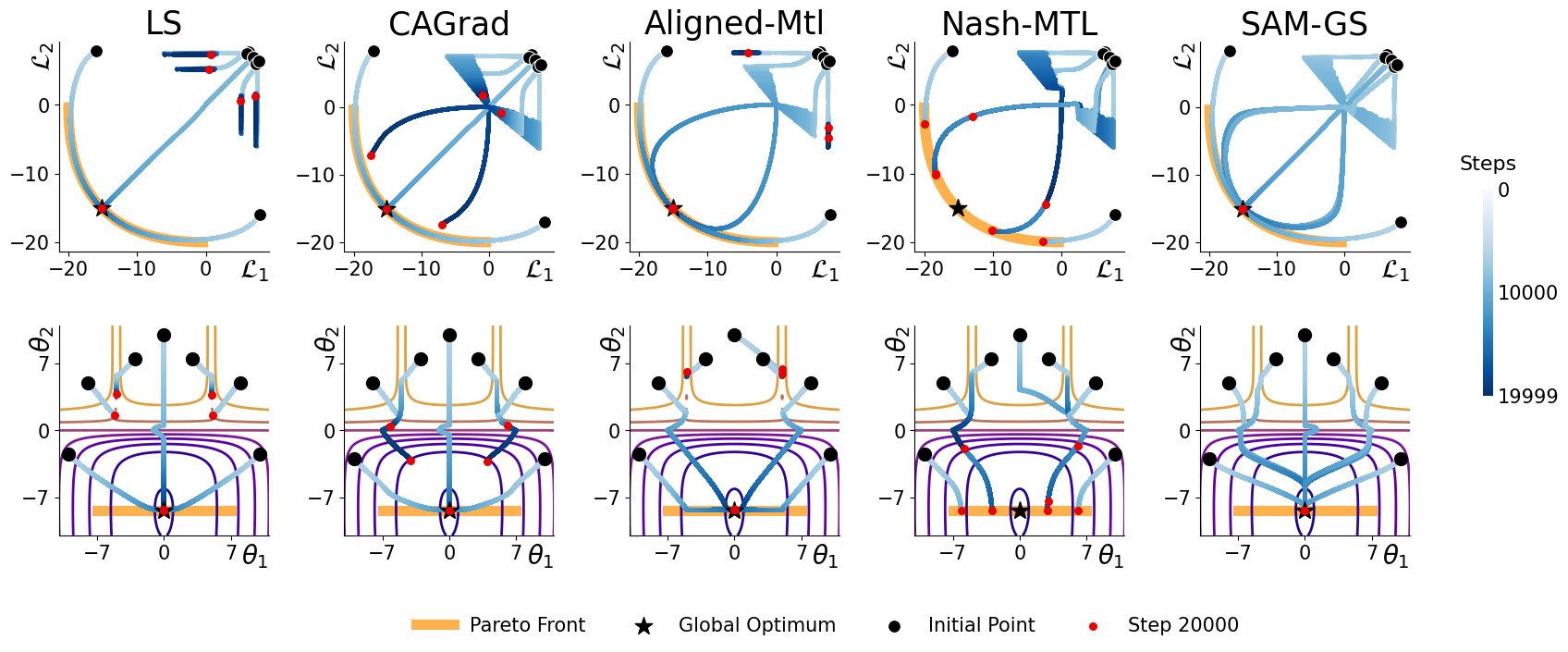}
    \caption{Trajectories for different methods starting from 7 different initial points: Linear Sum (LS) approach using Adam \cite{kingma2014adam}, Nash-MTL \cite{Navon2022_nashmtl}, CAGrad \cite{Liu2021NeurIPS_cagrad}, Aligned-MTL \cite{Senushkin2023_alighedmtl}, and SAM-GS, from the starting points to the global optimum at the centre of the Pareto front in the loss space (top row) and parameter space (bottom row). The red dots show the end state of the trajectory after 20,000 iterations.}
    \label{fig:toy_problem}
\end{figure}
In the experimental results shown in Figure \ref{fig:toy_problem}, indicate that the proposed approach exhibits behaviour comparable to CAGrad \cite{Liu2021NeurIPS_cagrad}. The maximum number of steps is set to 20,000: SAM-GS converges within 18,000 steps, and the simulation was run for 10\% more steps to ensure a good comparison. Our method is the only one that consistently reaches the global optimum from all the considered starting points.
This superior performance highlights the effectiveness of SAM-GS in navigating complex loss scenarios over existing methods.

\subsubsection{Two-task problem with two global optima.}\label{section:2-Task 2 Global Optimum}
\begin{figure}[t]
    \centering
    \begin{subfigure}{0.23\textwidth}
    \centering
    \includegraphics[width=\textwidth]{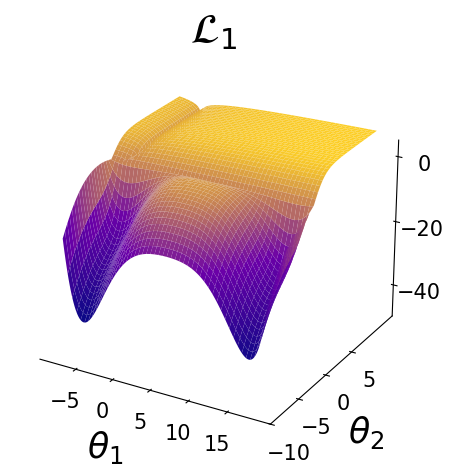}
        \caption{}
    \end{subfigure}
    \begin{subfigure}{0.23\textwidth}
    \centering
     \includegraphics[width=\textwidth]{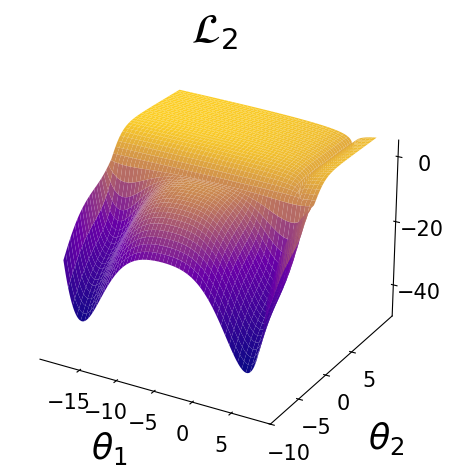}
        \caption{}
    \end{subfigure}
   \begin{subfigure}{0.23\textwidth}
   \centering
    \includegraphics[width=\textwidth]{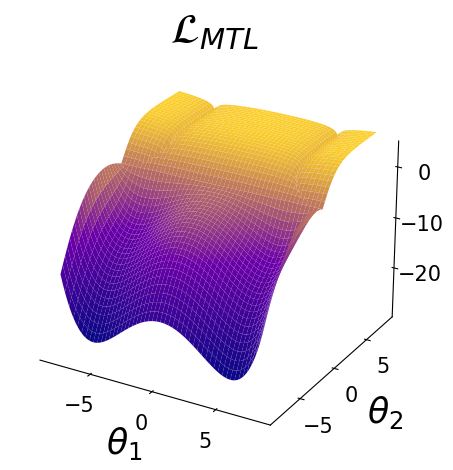}
        \caption{}
    \end{subfigure}
    \begin{subfigure}{0.23\textwidth}
   \centering
    \includegraphics[width=\textwidth]{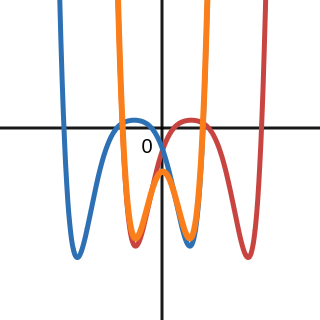}
        \caption{}
    \end{subfigure}
    \caption{Illustration of the multi-task optimisation problem ($\mathcal{L}_{MTL}$) computed as the sum of $\mathcal{L}_{1}$ and $\mathcal{L}_{2}$. In panel (d), the loss functions, $\mathcal{L}_{1}$ and $\mathcal{L}_{2}$ are displayed in red and blue, respectively, as a function of $\theta_1$ given $\theta_2 = -5$, and $\mathcal{L}_{MTL}$ is displayed in orange.}
    \label{fig:toy_3d_2o}
\end{figure}

We propose a novel inspired by Nash-MTL \cite{Navon2022_nashmtl}, where we introduce two distinct global optima to evaluate the MTL optimisation methods in a multi-optima scenario.
In this setup, two loss functions, each dependent on two parameters, exhibit one global optimum and one local optimum. The combination of these functions forms a multi-task optimisation problem with two global optima corresponding to the two local optima of the single task problems, as illustrated in Figure \ref{fig:toy_3d_2o}. This problem setup is interesting because the MTL optima correspond to the single-task local minima, thus challenging the optimisation process.
Additionally, this setup presents a saddle point, which is absent in the first synthetic problem, introducing a further complexity. The complete formulation of this setup is detailed in the supplementary material.

We compare our SAM-GS with LS using Adam \cite{kingma2014adam}, Nash-MTL \cite{Navon2022_nashmtl}, CAGrad \cite{Liu2021NeurIPS_cagrad}, and Aligned-MTL \cite{Senushkin2023_alighedmtl} across six different initialisation points, running the algorithm for a maximum of 20,000 steps.

\begin{figure}[h!t]
    \centering
    \includegraphics[width=\linewidth]{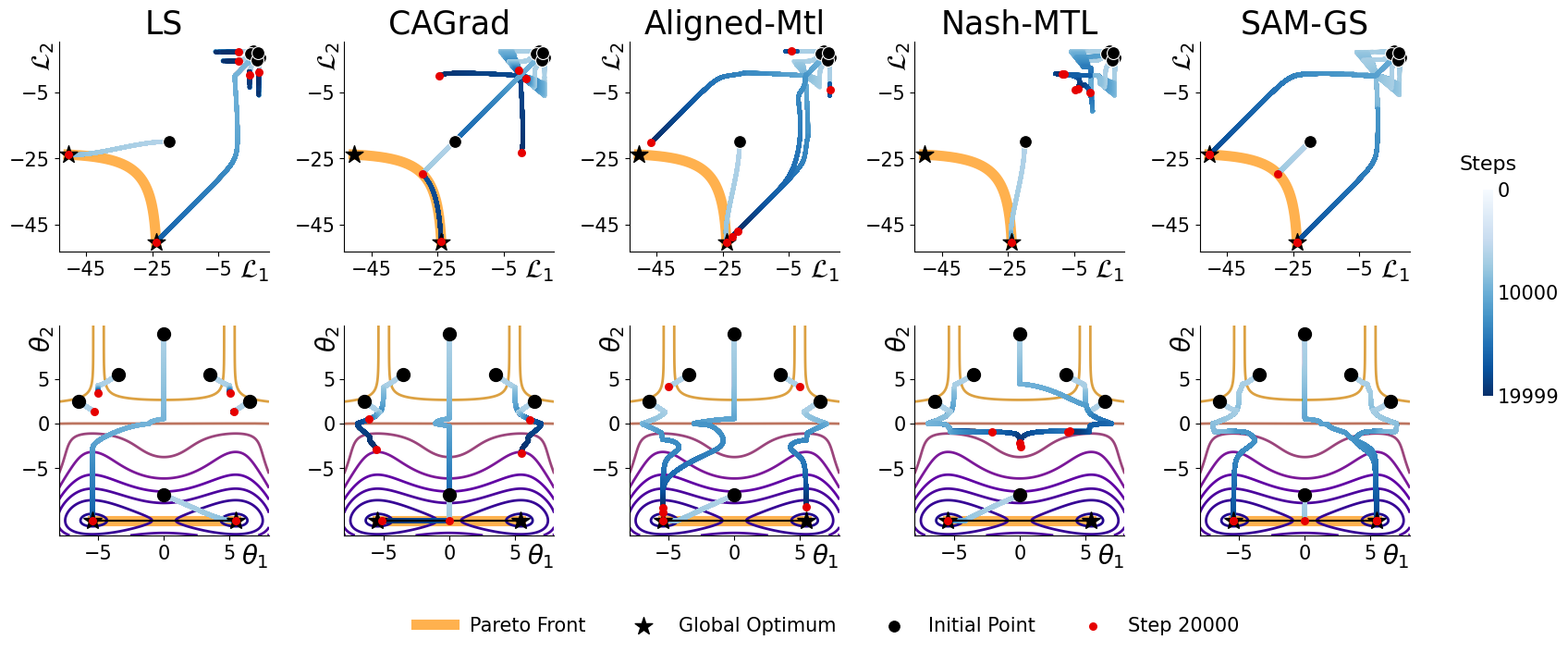}
    \caption{Trajectories for different methods in the second synthetic problem: Linear Sum (LS) approach using Adam \cite{kingma2014adam}, Nash-MTL \cite{Navon2022_nashmtl}, CAGrad \cite{Liu2021NeurIPS_cagrad}, Aligned-MTL \cite{Senushkin2023_alighedmtl}, and SAM-GS, starting from six initial points and converging to one global optima at the extremes of the Pareto front in the loss space (top row) and parameter space (bottom row).}
    \label{fig:toy_problem_2o}
\end{figure}

As shown in Figure \ref{fig:toy_problem_2o}, SAM-GS is the method that reaches one of the two global optima for most of the considered initial points within the maximum number of iterations.
The ability of the method to consistently and efficiently identify a global optimum across different initialisations highlights its potential for solving complex multi-task optimisation problems with multiple optima, ensuring faster and more reliable convergence than existing approaches.

\subsection{Performance Analysis}
We tested the effectiveness of SAM-GS on three different multi-task supervised benchmarks, which have been used by various competitive optimisation methods \cite{Liu2023NeurIPS_famo,Liu2021ICLR_imtl,Navon2022_nashmtl,Senushkin2023_alighedmtl}, CelebA \cite{Ziwei2015ICCV_celeba} (40 tasks), NYU-v2 \cite{Silberman2012ECCV_nyuv2} (3 tasks) and CityScapes \cite{Cordts2016CVPR_cityscapes} (2 tasks).
We compare SAM-GS against 14 different optimisation methods for multi-task learning. 
These include loss balancing methods such as UW \cite{Kendall2018CVPR_uw}, DWA \cite{Liu2019CVPR_dwa}, GradNorm \cite{Chen2018ICML_gradnorm}, and RGW \cite{Lin2022TRML_RGW}, as well as FAMO \cite{Liu2023NeurIPS_famo}. 
Additionally, we evaluate gradient surgery methods, including PCGrad \cite{Yu2020NeurIPS_pcgrad}, CAGrad \cite{Liu2021NeurIPS_cagrad}, GradDrop \cite{Chen2020NeurIPS_graddrop},MGDA \cite{Dong2015_mgda2}, IMTL \cite{Liu2021ICLR_imtl}, Nash-MTL \cite{Navon2022_nashmtl} and Aligned-MTL \cite{Senushkin2023_alighedmtl}. 

In the remainder of this section we present the evaluation metrics used for the comparative analysis, the results on three benchmarks and, finally, the ablation study on SAM-GS hyperparameter.

\subsubsection{Evaluation Metrics.}
To evaluate the performance of the optimisation methods, we use the Mean Ranking (\textbf{MR}) and the $\mathbf{\Delta m \%}$ metrics, similar to Nash-MTL \cite{Navon2022_nashmtl}. The $MR$ metric is the average rank of each method across tasks, where an $MR$ of $1$ indicates that the method ranks first on all tasks. The $\mathbf{\Delta m \%}$ metric, defined in Equation \eqref{formula:deltam}, quantifies the percentage improvement or degradation in performance of a method compared to the baseline single-task models.
\begin{equation}
    \Delta m \% = \frac{1}{K} \sum_{k=1}^K (-1)^{\nu_k} \frac{m_{mtl,k}-m_{stl,k}}{m_{stl,k}} \cdot 100
    \label{formula:deltam}
\end{equation}
Here, $m_{mtl,k}$ and $m_{stl,k}$ represent the performance metrics for the $MTL$ optimisation method and single-task models, respectively, for task $k.$ The binary indicator $\nu_k$ is set to 1 when a higher value of $m$ indicates better performance (e.g., accuracy), and $0$ when a lower value is preferable (e.g., error).

\subsubsection{CityScapes (2 tasks).}
The CityScapes dataset \cite{Cordts2016CVPR_cityscapes} contains $5,000$ street-level RGBD images with per-pixel annotations across $19$ semantic segmentation categories, grouped into $7$ main categories.
We adopt a similar experimental setup used in Nash-MTL \cite{Navon2022_nashmtl}, training a single Multi-Task Attention Network (MTAN) \cite{Liu2019CVPR_dwa} model to simultaneously perform depth estimation and semantic segmentation.
We identify that the best hyperparameter for SAM-GS are, $\beta_1 = 0.9$, $\beta_2=0.9$, $\gamma= 0.9$.
Results in Table \ref{tab:resultsCityScapes} shows that in this settings SAM-GS it is competitive with other methodology, but not superior in term of $\Delta m \%$. 
Some methods (e.g. UW \cite{Kendall2018CVPR_uw}) have strictly better $\Delta m \%$ by excelling in one task; our approach has more balanced competitive performance across all tasks.
The superior performance of Aligned-MTL \cite{Senushkin2023_alighedmtl}, which focuses only on \textit{angle-based gradient conflicts}, indicates that in this dataset inter-sample conflicts are more relevant, as also shown in \cite{Elich_2025_gcpr}. 
This may explain the limitations of the proposed approach for this dataset.

\begin{table}[h!t]
    \centering
    \setlength{\tabcolsep}{5pt}
    \caption{CityScapes results}
    \label{tab:resultsCityScapes}
    \resizebox{0.75\textwidth}{!}{ 
\begin{tabular}{lllllllll}
\toprule
         & \multicolumn{2}{c}{Segmentation} &  & \multicolumn{2}{c}{Depth} &  &      &        \\ \cline{2-3} \cline{5-6}
         & mIoU  {\small $\scriptstyle\uparrow$} & PixAcc {\small $\scriptstyle\uparrow$} &  & AbsErr {\small $\scriptstyle\downarrow$}  & RelErr {\small $\scriptstyle\downarrow$}  &  & \textbf{MR} {\small $\scriptstyle\downarrow$}     & \textbf{$\Delta m \%$} {\small $\scriptstyle\downarrow$}   \\ \midrule
STL      & 74.01          & 93.16           &  & 0.0125     & 27.77        &  &      &        \\ \hline
LS       & 71.0           & 91.7            &  & 0.0161     & 33.8         &  & 11.8 & 14.1   \\
SI       & 71.0           & 91.7            &  & 0.0161     & 33.8         &  & 11.8 & 14.1   \\
RLW      & 74.6           & 93.4            &  & 0.0158     & 47.8         &  & 11.0 & 24.4   \\
DWA      & 75.2           & 93.5            &  & 0.016      & 44.4         &  & 8.5  & 21.4   \\
UW       & 72.0           & 92.8            &  & 0.014      & \textbf{30.1}&  & 7.75 & 5.89   \\
MGDA     & 68.8           & 91.5            &  & 0.0309     & 33.5         &  & 12.5 & 44.1   \\
PCGrad   & 75.1           & 93.5            &  & 0.0154     & 42.1         &  & 9.12 & 18.3   \\
GradNorm & 73.7           & 93.0            &  & 0.0124     & 34.1         &  & 7.75 & 5.63   \\
GradDrop & 75.3           & 93.5            &  & 0.0157     & 47.5         &  & 7.75 & 23.7   \\
CAGrad   & 75.2           & 93.5            &  & 0.0141     & 37.6         &  & 7.88 & 11.6   \\
IMTL-G   & 75.3           & 93.5            &  & 0.0135     & 38.4         &  & 6    & 11.1   \\
Nash-MTL  & 75.4  & \textbf{93.7}   &  & \textbf{0.0129}     & 35.0        &  & 3.75 & 6.82   \\
FAMO     & 74.5           & 93.3            &  & 0.0145     & 32.6         &  & 7.50 & 8.13   \\ 
Aligned-MTL&\textbf{75.8}& 93.7           &  & 0.0133     & 32.66        &  & \textbf{2}     & \textbf{5.27} \\
\midrule
SAM-GS   & 75.2           & 93.5            &  & 0.0136     & 33.1         &  & 5.00 & 6.41  \\ 
\bottomrule
\end{tabular}
}
\end{table}

\subsubsection{NYU-V2 (3 tasks).}
The NYU-v2 dataset \cite{Silberman2012ECCV_nyuv2} comprises $1,449$ RGBD images of indoor scenes, with dense pixel-level annotations across $13$ classes. 
We follow a similar experimental setup to Nash-MTL \cite{Navon2022_nashmtl}, training a single MTAN \cite{Liu2019CVPR_dwa} model to perform depth estimation, image segmentation, and surface normal prediction.
We identify the following hyperparameters for SAM-GS: $\beta_1 = 0.9$, $\beta_2=0.9$, $\gamma= 0.9$.
The results in Table \ref{tab:resultsNYUv2} show superior performance of SAM-GS, compared to other methods, in cases with more than two tasks.

\begin{table}[h!t]
\centering
    \caption{NYU-V2 results}
    \label{tab:resultsNYUv2}
    \resizebox{\textwidth}{!}{ 
\begin{tabular}{llllllllllllllll}
\toprule
         & \multicolumn{2}{c}{Segmentation} &  & \multicolumn{2}{c}{Depth} &  & \multicolumn{6}{c}{Surface Normal}       &  &       &       \\ \cline{2-3} \cline{5-6} \cline{8-13}
 &
  \multicolumn{1}{c}{mIoU{\small $\scriptstyle\uparrow$}} &
  Pix Acc {\small $\scriptstyle\uparrow$} &
   &
  \multicolumn{1}{c}{Abs Err {\small $\scriptstyle\downarrow$}} &
  \multicolumn{1}{c}{Rel Err {\small $\scriptstyle\downarrow$}} &
   &
  \multicolumn{2}{c}{Angle Dist {\small $\scriptstyle\downarrow$}} &
   &
  \multicolumn{3}{c}{Within t° {\small $\scriptstyle\uparrow$}} &
   &
   &
   \\ \cline{8-9} \cline{11-13}
 &
  \multicolumn{1}{c}{} &
  \multicolumn{1}{c}{} &
   &
  \multicolumn{1}{c}{} &
  \multicolumn{1}{c}{} &
   &
  \multicolumn{1}{c}{Mean} &
  \multicolumn{1}{c}{Median} &
   &
  \multicolumn{1}{c}{11.25} &
  \multicolumn{1}{c}{22.5} &
  \multicolumn{1}{c}{30} &
   &
  \multicolumn{1}{c}{\textbf{MR} {\small $\scriptstyle\downarrow$}} &
  \multicolumn{1}{c}{\textbf{$\Delta m \%$} {\small $\scriptstyle\downarrow$}} \\ \midrule
STL      & 38.3            & 63.76          &  & 0.6754      & 0.278       &  & 25.01 & 19.21 &  & 30.14 & 57.2  & 69.15 &  &       &      \\ \midrule
LS       & 39.29           & 65.33          &  & 0.5493      & 0.2263      &  & 28.15 & 23.96 &  & 22.09 & 47.5  & 61.08 &  & 11.4  & 5.59  \\
SI       & 38.45           & 64.27          &  & 0.5354      & 0.2201      &  & 27.6  & 23.37 &  & 22.53 & 48.57 & 62.32 &  & 10.3  & 4.39  \\
RLW      & 37.17           & 63.77          &  & 0.5759      & 0.241       &  & 28.27 & 24.18 &  & 22.26 & 47.05 & 60.62 &  & 13.8 & 7.78  \\
DWA      & 39.11           & 65.31          &  & 0.551       & 0.2285      &  & 27.61 & 23.18 &  & 24.17 & 50.18 & 62.39 &  & 10.2  & 3.57  \\
UW       & 36.87           & 63.17          &  & 0.5446      & 0.226       &  & 27.04 & 22.61 &  & 23.54 & 49.05 & 63.65 &  & 10.0  & 4.05  \\
MGDA     & 30.47           & 59.9           &  & 0.607       & 0.2555      &  & 24.88 & 19.45 &  & 29.18 & 56.88 & 69.36 &  & 7.4     & 1.38  \\
PCGRAD   & 38.06           & 64.64          &  & 0.555       & 0.2325      &  & 27.41 & 22.8  &  & 23.86 & 49.83 & 63.14 &  & 10.6     & 3.97  \\
GradNorm & 20.09           & 64.64          &  & 0.7200      & 0.2800      &  & \textbf{24.83} & \textbf{18.86} &  & \textbf{30.81} & \textbf{57.94} & \textbf{69.73} & & 7.2 & 7.22 \\
GradDrop & 39.39           & 65.12          &  & 0.5455      & 0.2279      &  & 27.48 & 22.96 &  & 23.38 & 49.44 & 62.87 &  & 9.6     & 3.58  \\
CAGrad   & 39.79           & 65.49          &  & 0.5486      & 0.225       &  & 26.31 & 21.58 &  & 25.61 & 52.36 & 65.58 &  & 7.1  & 0.2   \\
IMTL-G   & 39.35           & 65.6           &  & 0.5426      & 0.2256      &  & 26.02 & 21.19 &  & 26.2  & 53.13 & 66.24 &  & 6.3  & -0.76 \\
Nash-MTL  & 40.13           & 65.93         &  & 0.5261      & 0.2171     &  & 25.26 & 20.08 &  & 28.4  & 55.47 & 68.15 &  & 4.2  & -4.04 \\
FAMO     & 38.88           & 64.9           &  & 0.5474      & 0.2194      &  & 25.06 & 19.57 &  & 29.21 & 56.61 & 68.98 &  & 4.8  & -4.1  \\ 
Aligned-MTL& \textbf{40.82}         & 66.33          &  & 0.5300      & 0.2200      &  & 25.19 & 19.71 &  & 28.88 & 56.23 & 68.54 &  & 3.6  & -4.93 \\ 
\midrule
SAM-GS    & 40.79           & \textbf{66.46}           &  &\textbf{ 0.5251}      & \textbf{0.2169}      &  & 25.03 & 19.65 &  & 29.26 & 56.35 & 68.78 &  & \textbf{2.4}  & \textbf{-5.3}\\  
\bottomrule
\end{tabular}
}
\end{table}

\subsubsection{CelebA (40 tasks).}
The CelebA dataset \cite{Ziwei2015ICCV_celeba} is a collection of $200,000$ facial images of $10,000$ distinct celebrities, with $40$ binary annotations of facial attributes for each image. 
We use the experimental setup outlined in FAMO \cite{Liu2023NeurIPS_famo}, training a CNN model to perform $40$ binary classification tasks.
The hyperparameter search on the validation data identifies the following as the best hyperparameters for SAM-GS: $\beta_1 = 0.9$, $\beta_2=0.99$, $\gamma= 0.9$.
The results in Table \ref{tab:resultsCeleba} show a superior performance of SAM-GS in handling $40$ different tasks concurrently.
\begin{table}[t]
    \centering
     \begin{minipage}{0.3\textwidth}
        \centering
        \caption{CelebA results.}
        \label{tab:resultsCeleba}
        \resizebox{\textwidth}{!}{ 
        \begin{tabular}{lll}
            \toprule
             \textbf{Method}  & \textbf{$\Delta m \%$} {\small $\scriptstyle\downarrow$}   \\ \hline
            LS          &  6.28         \\
            SI          &  7.83         \\
            RLW         &  5.22         \\
            DWA         &  6.95         \\
            UW          &  5.78         \\
            MGDA        &  10.93        \\
            PCGrad      &  6.65         \\
            GradDrop    &  7.80         \\
            CAGrad      &  6.20         \\
            IMTL-G      &  4.67         \\
            Nash-MTL    &  4.97         \\
            FAMO        &  4.72         \\ 
            Aligned-MTL &  4.58         \\
            \midrule
            SAM-GS      & \textbf{3.33} \\ 
            \bottomrule
        \end{tabular}
        }
    \end{minipage}
    \hspace{0.2cm}
    \begin{minipage}{0.55\textwidth}
        \centering
        \caption{Reinforcement learning (MT10).}
        \label{tab:rl_mt10}
        \resizebox{\textwidth}{!}{
        \begin{tabular}{lc}
            \toprule
            \textbf{Method}   & \textbf{Success $(mean \pm stderr)$} \\ 
            \midrule
            STL SAC           & $0.90 \pm 0.032$       \\ \midrule
            MTL SAC           & $0.49 \pm 0.073$       \\
            MTL SAC + TE      & $0.54 \pm 0.047$       \\
            MH SAC            & $0.61 \pm 0.036$       \\
            SM                & $0.73 \pm 0.043$       \\
            CARE              & $0.84 \pm 0.051$       \\
            PCGrad            & $0.72 \pm 0.022$       \\
            CAGrad            & $0.83 \pm 0.045$       \\
            Nash-MTL          & $0.91 \pm 0.031$       \\
            Aligned-MTL       & \textbf{0.97 $\pm$ 0.045}       \\ 
            FAMO              & $0.83 \pm 0.05$        \\ \midrule
            SAM-GS            & $0.91 \pm  0.018 $     \\
            \bottomrule
        \end{tabular}
        }
    \end{minipage}
\end{table}
\subsubsection{Ablation study on $\gamma.$}\label{section:ablation}
In this section we provide a systematic study over the values of the similarity threshold $\gamma$.

\begin{figure}[h!t]
    \centering
    \includegraphics[width=\textwidth]{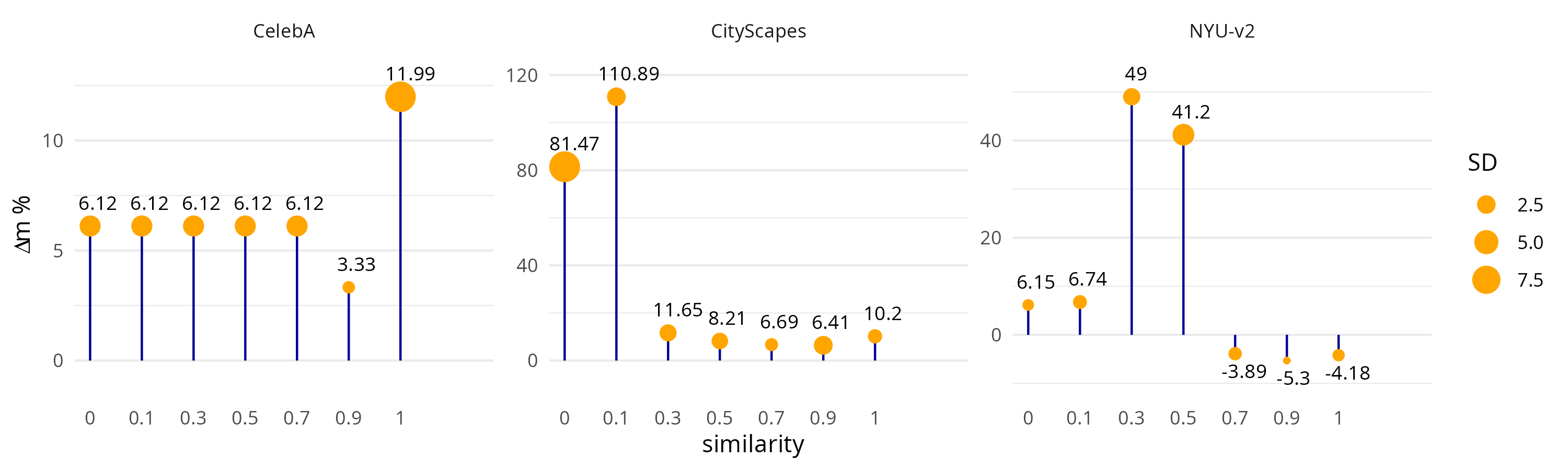}
    \caption{Ablation study over $\gamma$. The plot shows the performance, in terms of $\Delta m \%$, of SAM-GS across three supervised learning settings with $\gamma$ values of $\{0, 0.1, 0.3, 0.5, 0.7, 0.9, 1\},$ including standard deviation (SD)}
    \label{fig:ablation}
\end{figure}
Figure \ref{fig:ablation} highlights the critical role of $\gamma$ in model performance. 
Extreme settings ($\gamma=0$ or $\gamma=1$), which make the algorithm to rely exclusively on either the equalisation or the momentum component of SAM-GS, yield suboptimal results. 
On the other hand, intermediate values of $\gamma$, with a general trend towards higher settings, yield preferable results.

\subsection{MTDL Reinforcement Learning (10 tasks)}
Finally, we tested SAM-GS on a multi-task reinforcement learning (RL) problem against the most relevant $MTL$ methods specifically designed for RL problems and a selection of the most recent gradient surgery methods. Specifically, we applied a variation of the SAM-GS method to the MetaWorld \cite{Tianhe_metaworld} MT10 benchmark, which comprises 10 distinct robot manipulation tasks with various reward functions. 
The variation concerns the computation of \( \Psi_t \); we found that using \( \Psi_t = \min \psi(g_i, g_j)\) led to improved results compared to averaging in a multi-task reinforcement learning problem. 
The experimental setting is similar to the one used in CAGrad \cite{Liu2021NeurIPS_cagrad}, using Soft Actor-Critic (SAC) \cite{Guo_2018_ECCV} as a baseline, trained with various gradient manipulation methods \cite{Yu2020NeurIPS_pcgrad,Liu2021NeurIPS_cagrad,Navon2022_nashmtl,Senushkin2023_alighedmtl,Liu2023NeurIPS_famo}.
We also evaluate MTL-RL \cite{Sodhani_2021_icml} approaches, including MTL SAC, Multi-task SAC with task encoder (MTL SAC + TE) \cite{Wulfmeier_2020_RSS}, Multi-headed SAC (MH SAC) \cite{Wulfmeier_2020_RSS}, Soft Modularization (SM) \cite{Ruihan_2020_NeurIPS}, and CARE \cite{Sodhani_2021_icml}.
We identify the hyperparameters for SAM-GS: $\beta_1 = 0.9$, $\beta_2=0.99$, $\gamma= 0.9$.
The results presented in Table \ref{tab:rl_mt10} indicate that SAM-GS achieves performance levels on par with Nash-MTL \cite{Navon2022_nashmtl}, while surpassing STL baseline, FAMO \cite{Liu2023NeurIPS_famo}, CAGrad \cite{Liu2021NeurIPS_cagrad}, and the standard gradient descent baseline method.

\section{Conclusions}\label{Sec:Conclusions}
In multi-task deep learning training a single model on many tasks can be affected by potentially conflicting task gradients that would hinder the concurrent convergence of the diverse loss functions. In this study, the importance of the gradient magnitude similarity for the effective overall optimisation of the model has been studied and highlighted. As a result, a novel gradient surgery method, the Similarity-Aware Momentum Gradient Surgery (SAM-GS), has been proposed. SAM-GS is based on a measure of the task gradient magnitude similarity and used to control and guide two mechanisms: a momentum-based regularisation and a remedy for gradient magnitude conflicts.
An extensive evaluation has demonstrated that SAM-GS effectively addresses a range of challenges with respect to task gradient conflicts and outperforms previous optimisation methods in two synthetic problems, several benchmarks from real-world computer vision applications, and a benchmark for reinforcement learning tasks.
Future work may include a theoretical analysis of convergence to provide optimisation guarantees.
Moreover, a direction for further improvements is the analysis of the current limitations to address strict stationary states such as saddle points, where task gradients have very similar magnitude and opposite directions.

\begin{credits}
\subsubsection{\ackname} We would like to thank the anonymous reviewers for their thorough reviews and insightful comments.
\end{credits}

\bibliographystyle{splncs04}
\bibliography{ref}

\newpage
\appendix
\section{The Synthetic Problem}
\subsection{Two-Task Problem with One Global Optimum}\label{APPENDIX:toy1globla}

The synthetic problem with one global optimum is formulated as reported in \cite{Liu2021NeurIPS_cagrad} where the two loss functions are equally weighted. 
In order to allow the replication the results presented in the main paper, here we provide the detailed configuration for the seven initial points: 
\[\theta_{init} \in \{(-8, 5.0),(-3, 7.5), (0, 10.0),(3, 7.5), (8, 5.0), (-10,-2.5), (10,-2.5)\}.\]
We use the $ADAM$ \cite{kingma2014adam} optimiser with a learning rate of $1e-3$ \cite{Liu2021NeurIPS_cagrad}.

Moreover, here we present further results with different weighting of the loss functions as real-world problems may require non-equal weights and to allow a comparison to the work in \cite{Liu2023NeurIPS_famo,Navon2022_nashmtl,Senushkin2023_alighedmtl}.

\begin{figure}[H]
    \centering
    \includegraphics[width=\linewidth]{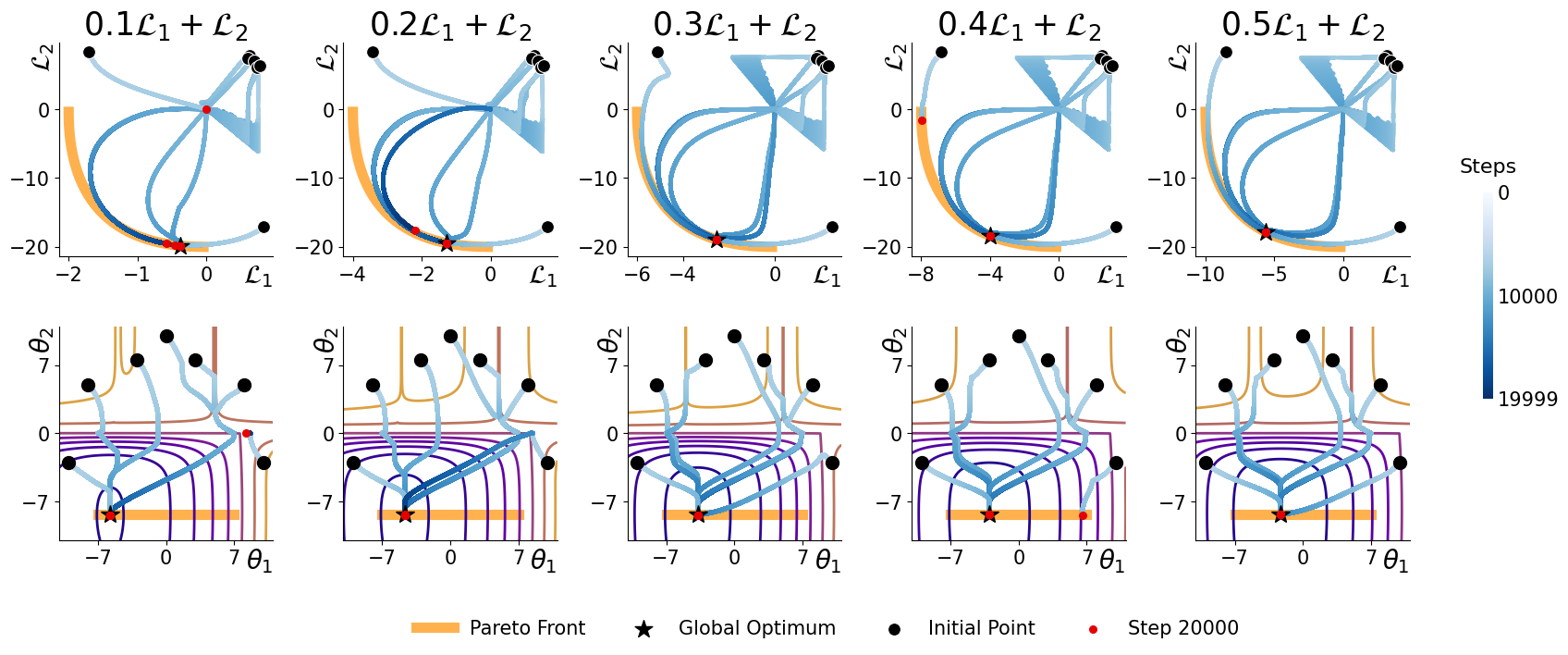}
    \caption{SAM-GS trajectories with different values of the loss weighting parameter.}
    \label{fig:ablation_pareto}
\end{figure}

We investigate a number of combinations of task weighting such that $\mathcal{L}_{mtl} = \alpha\mathcal{L}_1 + \mathcal{L}_2$.
The results in Figure \ref{fig:ablation_pareto} illustrate the different optimisation trajectories of SAM-GS using the same hyperparameters ($\beta_1 = 0.9$, $\beta_2 = 0.9$, $\gamma=0.1$) over the same number of steps (20000). The results show the strong performance of SAM-GS also in these scenarios with varying loss function weighting. 

\subsection{Two-Task Problem with Two Global Optima}\label{APPENDIX:toy2globla}
The synthetic problem with two global optima consists of two task objectives, each exhibiting a local optimum and a global optimum. The combination of these task objectives results in a function with two distinct global optima, providing a problem that is useful to analyse optimisation dynamics in the presence of multiple optima.
The model parameters are $\theta = (\theta_1, \theta_2) \in \mathbb{R}^2$ and the task objectives are $L_1$ and $L_2$.

\begin{align*}
    L_1(\theta) &= c_1(\theta) f_1(\theta) + c_2(\theta) g_1(\theta)\\
    L_2(\theta) &= c_1(\theta) f_2(\theta) + c_2(\theta) g_2(\theta)
\end{align*}

where

\begin{align*}
    f_1(\theta) &= \log \left( \max\left(\left| 0.5(-\theta_1 - 7) - \tanh(-\theta_2) \right|, 0.000005\right) \right) + 6, \\
    f_2(\theta) &= \log \left( \max\left(\left| 0.5(-\theta_1 + 3) - \tanh(-\theta_2) + 2 \right|, 0.000005\right) \right) + 6, \\
    c_1(\theta) &= \max(\tanh(0.5 \cdot \theta_2), 0), \\
    c_2(\theta) &= \max(\tanh(-0.5 \cdot \theta_2), 0),\\
    g_1(\theta) &= 0.1 \sum_{i=1}^{2} \left( \frac{\theta_i - d_{1i}}{4} \right)^6 - \sum_{i=1}^{2} \left( \frac{\theta_i - d_{2i}}{4} \right)^4 - 1.5 \sum_{i=1}^{2} \left( \frac{\theta_i - d_{2i}}{4} \right)^2 + 1.5,\\
    g_2(\theta) &= 0.1 \sum_{i=1}^{2} \left( \frac{\theta_i + d_{1i}}{4} \right)^6 - \sum_{i=1}^{2} \left( \frac{\theta_i + d_{2i}}{4} \right)^4 - 1.5 \sum_{i=1}^{2} \left( \frac{\theta_i + d_{2i}}{4} \right)^2 + 1.5,
\end{align*}

where $ d_{1} = (5.45, 0)$, $d_{2} = (5.5, 0)$

We set the weighting parameter $\alpha$ to 1 and use seven initial points:
\[\theta_{init}  \in \{(-3.5, 5.5),(3.5, 5.5), (-6.5, 2.5), (6.5, 2.5), (0, 10), (0, -8)\}.\]
We run the algorithm with the optimiser $ADAM$ \cite{kingma2014adam} with a learning rate of $1e-3$ and for 20,000 steps. The results are reported in Figure 4 of the main paper.

\begin{figure}[H]
    \centering
    \includegraphics[width=\linewidth]{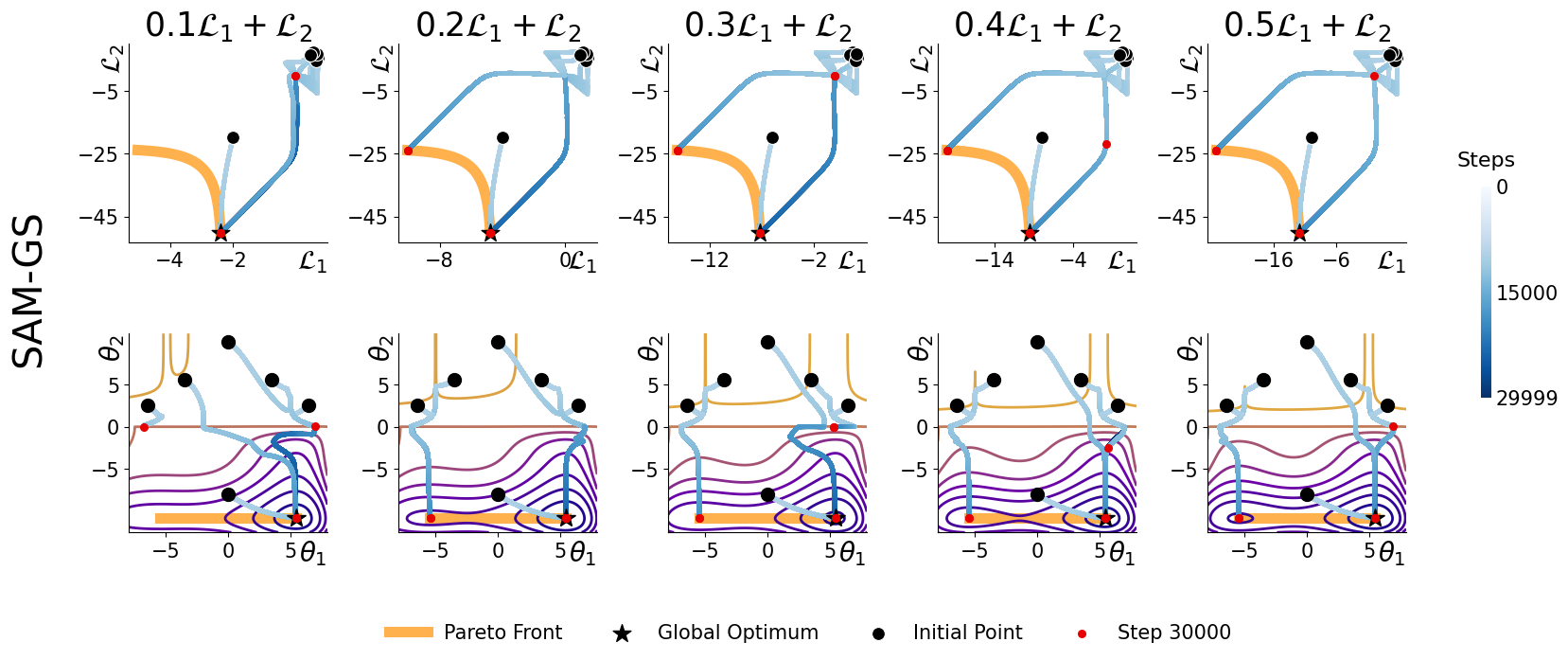}
    \caption{trajectories with different values of the loss weighting parameter on the second synthetic problem.}
    \label{ablation_2pareto}
\end{figure}

\begin{figure}[H]
    \centering
    \includegraphics[width=\textwidth]{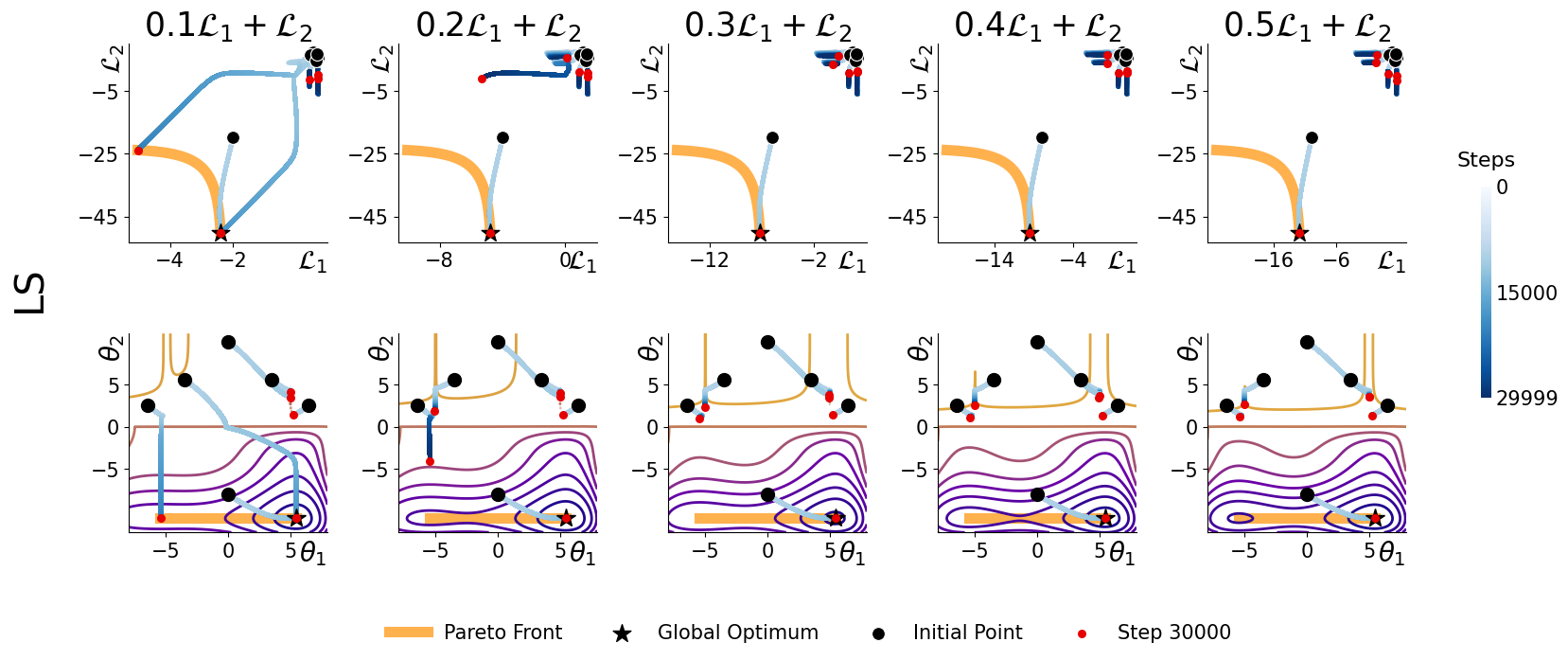}\\
    \includegraphics[width=\textwidth]{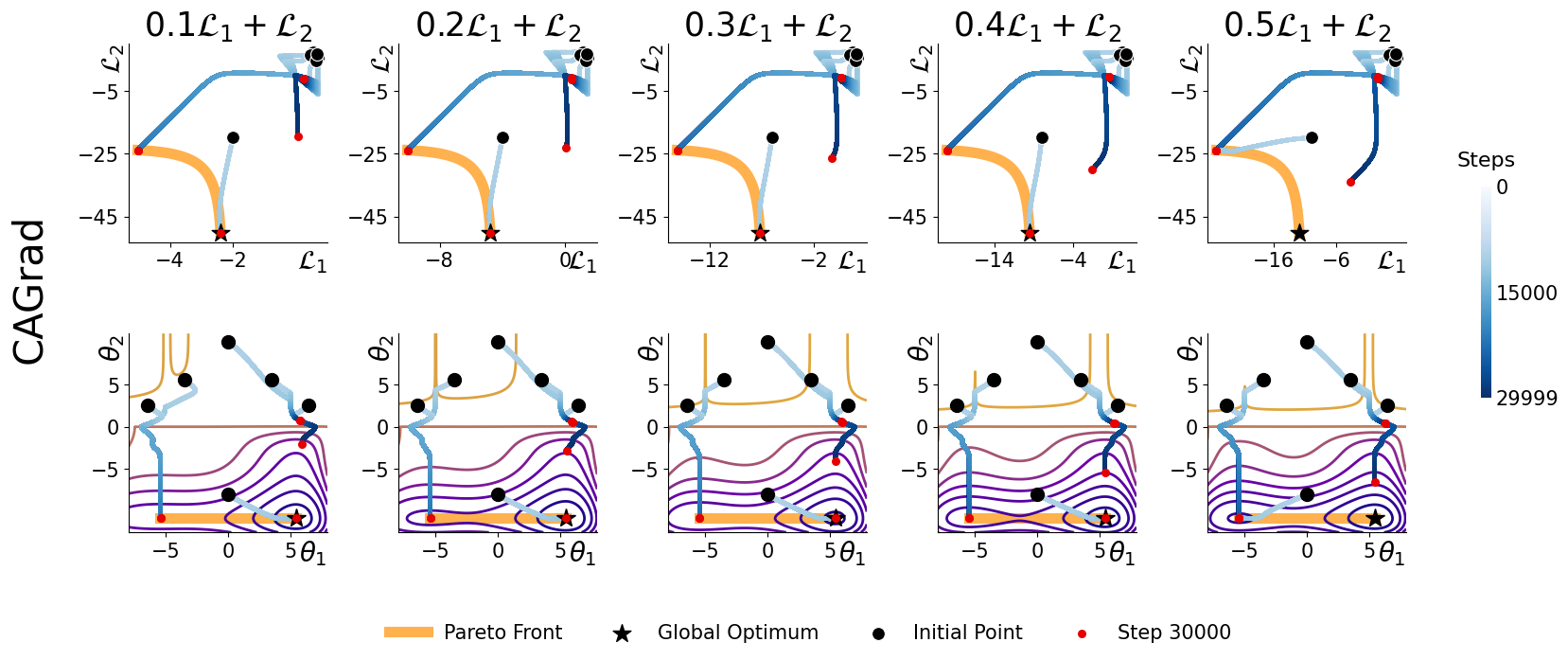}\\
    \includegraphics[width=\textwidth]{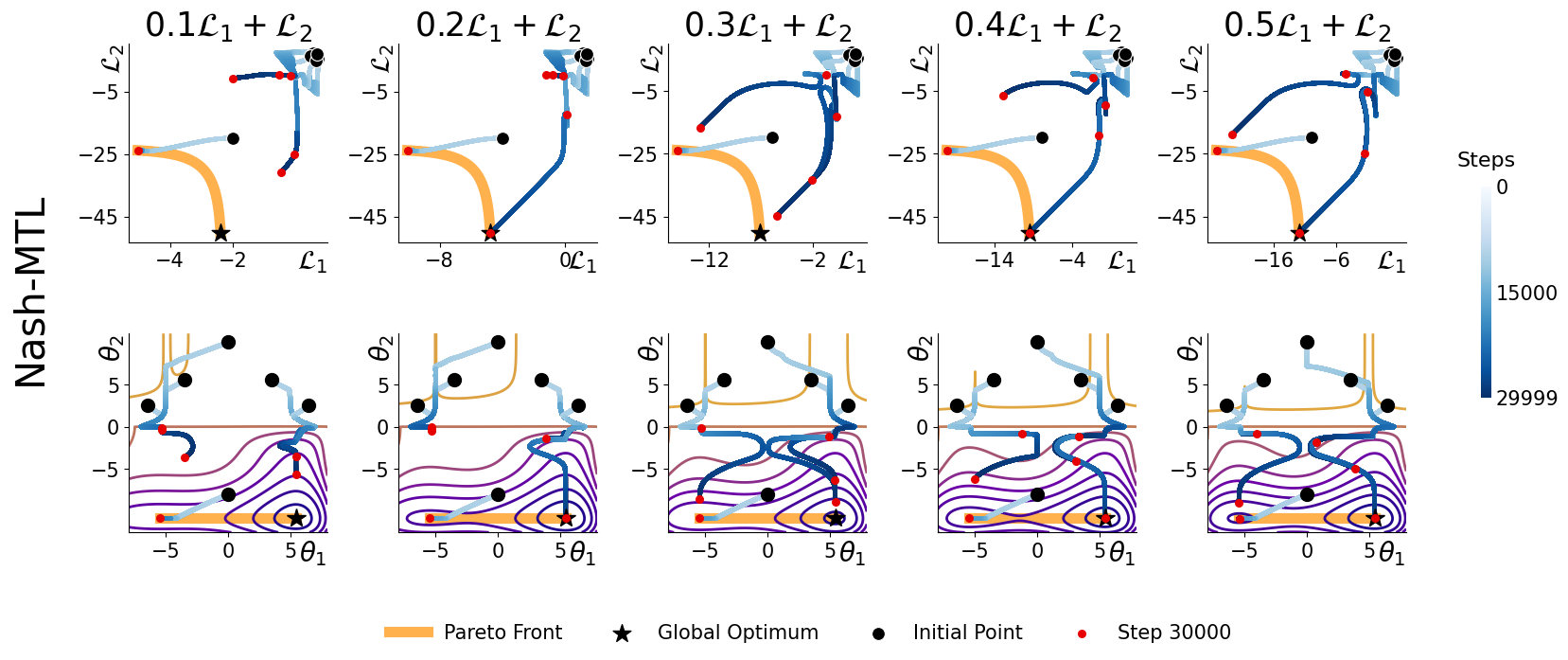}\\
    \includegraphics[width=\textwidth]{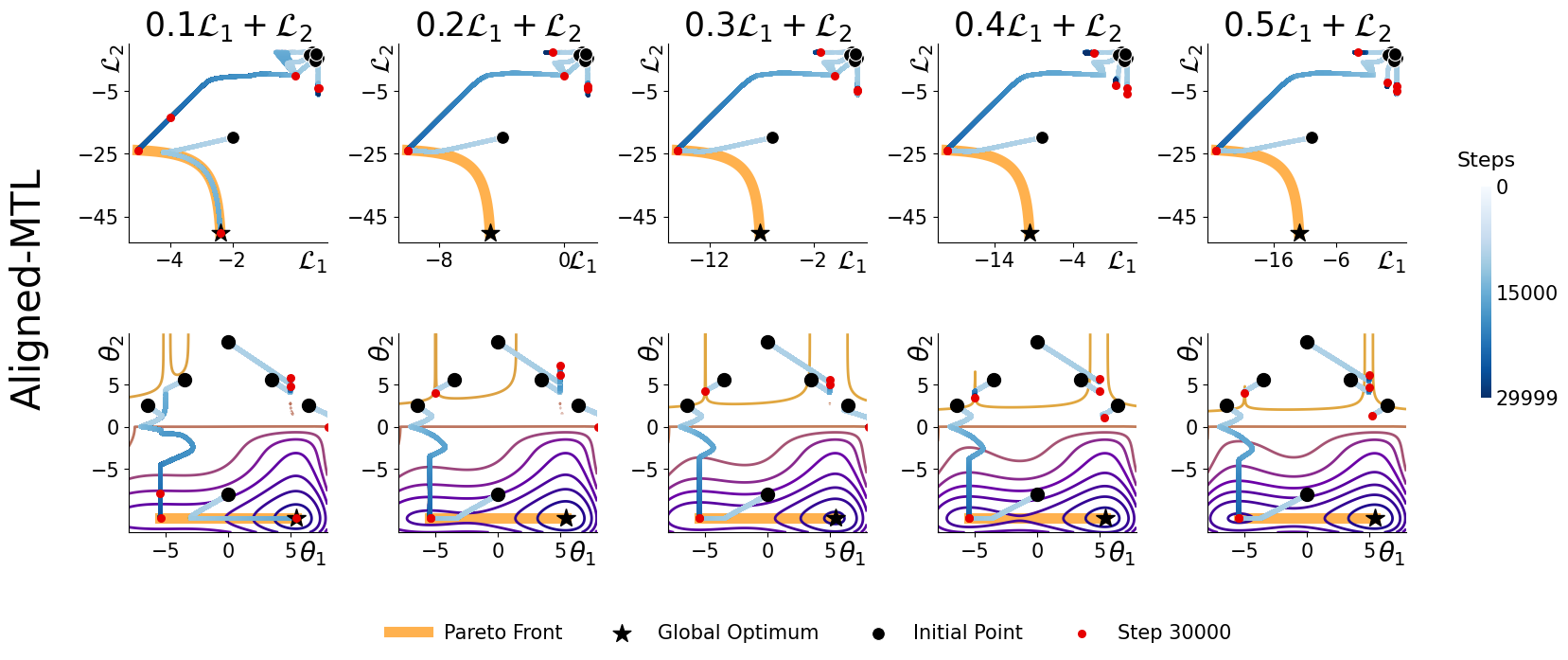}
    \caption{MTL optimisation methods trajectories with different values of the loss weighting parameter.}
    \label{ablation_2paretoM}
\end{figure}

We hereby present additional results in which the task weightings vary. In particular, we investigate multiple combinations of task weights, such that $\mathcal{L}_{mtl} = \alpha\mathcal{L}_1 + \mathcal{L}_2$.
The variation of the task weights is of particular interest, as it results in a transformation of the task landscape, thereby shifting the problem from a two global optima to a one local optimum and one global optimum problem setting. 
We evaluated different optimisation methods over 30,000 steps. The results, presented in Figure \ref{ablation_2pareto}, highlight the challenges of this problem and demonstrate the strong performance of SAM-GS in the proposed synthetic task. 
For comparison, we also include a similar analysis for other $MTL$ optimisation methods in Figure \ref{ablation_2paretoM}. 
SAM-GS performs slightly better than the other methodologies, consistently demonstrating its advantages. 

\section{Experimental Settings for the Performance Analysis of sec 5.2}

\subsection{CityScapes}
The model is trained for 200 epochs with a batch size of 8, using the Adam optimiser \cite{kingma2014adam} and a learning rate of $1e-4,$ and a learning rate scheduler that reduces the learning rate by a factor of $0.5$ every $100$ steps.
We tested $5$ different values of $\gamma$, $[0, 0.5, 0.7, 0.9, 1]$ and $3 $different values of $\beta_2,$ $[0.9, 0.95, 0.99].$
Table \ref{tab:resultsCityScapesError} provides the results from Section 5.2 with the additional information of the standard error to facilitate future comparisons.

\begin{table}[h!t]
    \centering
    \setlength{\tabcolsep}{5pt}
    \caption{CityScapes results}
    \label{tab:resultsCityScapesError}
    \resizebox{0.8\textwidth}{!}{ 
\begin{tabular}{lllllllll}
\toprule
         & \multicolumn{2}{c}{Segmentation} &  & \multicolumn{2}{c}{Depth} &  &      &        \\ \cline{2-3} \cline{5-6}
         & mIoU  {\small $\scriptstyle\uparrow$} & PixAcc {\small $\scriptstyle\uparrow$} &  & AbsErr {\small $\scriptstyle\downarrow$}  & RelErr {\small $\scriptstyle\downarrow$}  &  & \textbf{$\Delta m \%$} {\small $\scriptstyle\downarrow$}   \\ \midrule
SAM-GS (mean)   & 75.2           & 93.5            &  & 0.0136     & 33.1         &  & 6.41  \\ 
SAM-GS (stderr) & $\pm$ 0.000442    & $\pm$ 0.000124    &  & $\pm$ 0.0000931 & $\pm$ 0.500   &  & $\pm$ 0.566 \\
\bottomrule
\end{tabular}             
}
\end{table}

\subsection{CelebA}
The model is trained for $15$ epochs with a batch size of 256, using the Adam optimiser \cite{kingma2014adam} and a learning rate of $3e-4.$
We evaluated five values of $\gamma$ $[0,0.5,0.7,0.9,1]$ and three values of $\beta_2$ $[0.9,0.95,0.99].$
The results reported in Section 5.2 are computed on the test set using the model that achieved the best validation performance, averaged over three random seeds. Table \ref{tab:resultscelebserror} provides the results from Section 5.2 with the additional information of the standard error for future comparisons.

\begin{table}[h!t]
\centering
    \caption{CelebA results}
    \label{tab:resultscelebserror}
\begin{tabular}{lll}
            \toprule
             \textbf{Method}  & \textbf{$\Delta m \%$} {\small $\scriptstyle\downarrow$}   \\ \midrule
            SAM-GS (mean)      & 3.33 \\ 
            SAM-GS (stderr)    & $\pm$ 0.940 \\
            \bottomrule
        \end{tabular}
\end{table}
\subsection{NYU-v2}
The model is trained for $200$ epochs with a batch size of $2,$ using the $ADAM$ optimiser \cite{kingma2014adam} and a learning rate of $1e-4,$ and a learning rate scheduler that reduces the learning rate by a factor of 0.5 every 100 steps.
We tested 5 different values of $\gamma$, [0, 0.5, 0.7, 0.9, 1] and 3 different values of $\beta_2$, [0.9, 0.95, 0.99].
The results reported in Section 5.2 are averaged over the last 10 epochs and three random seeds. Table \ref{tab:resultsNYUv2error} provides the results from Section 5.2 with the additional information of the standard error for future comparisons.

\begin{table}[h!t]
\centering
    \caption{NYU-V2 results}
    \label{tab:resultsNYUv2error}
    \resizebox{\textwidth}{!}{ 
\begin{tabular}{llllllllllllllll}
\toprule
         & \multicolumn{2}{c}{Segmentation} &  & \multicolumn{2}{c}{Depth} &  & \multicolumn{6}{c}{Surface Normal}       &  &       &       \\ \cline{2-3} \cline{5-6} \cline{8-13}
 &
  \multicolumn{1}{c}{mIoU{\small $\scriptstyle\uparrow$}} &
  Pix Acc {\small $\scriptstyle\uparrow$} &
   &
  \multicolumn{1}{c}{Abs Err {\small $\scriptstyle\downarrow$}} &
  \multicolumn{1}{c}{Rel Err {\small $\scriptstyle\downarrow$}} &
   &
  \multicolumn{2}{c}{Angle Dist {\small $\scriptstyle\downarrow$}} &
   &
  \multicolumn{3}{c}{Within t° {\small $\scriptstyle\uparrow$}} &
   &
   &
   \\ \cline{8-9} \cline{11-13}
 &
  \multicolumn{1}{c}{} &
  \multicolumn{1}{c}{} &
   &
  \multicolumn{1}{c}{} &
  \multicolumn{1}{c}{} &
   &
  \multicolumn{1}{c}{Mean} &
  \multicolumn{1}{c}{Median} &
   &
  \multicolumn{1}{c}{11.25} &
  \multicolumn{1}{c}{22.5} &
  \multicolumn{1}{c}{30} &
   &
  \multicolumn{1}{c}{\textbf{$\Delta m \%$} {\small $\scriptstyle\downarrow$}} \\ \midrule
SAM-GS (mean)   & 40.79     & 66.46      &  & 0.5251    & 0.2169    &  & 25.03     & 19.65     &  & 29.26     & 56.35     & 68.78     &  & -5.3\\  
SAM-GS (stderr) & $\pm$ 0.172 & $\pm$ 0.104  &  & $\pm$ 0.003 & $\pm$ 0.002 &  & $\pm$ 0.029 & $\pm$ 0.066 &  & $\pm$ 0.144 & $\pm$ 0.135 & $\pm$ 0.094 &  & $\pm$ 0.147 \\ 
\bottomrule
\end{tabular}
}
\end{table}

\end{document}